\documentclass[conference]{IEEEtran}
\usepackage{times}

\usepackage[numbers]{natbib}
\usepackage{multicol}
\usepackage[bookmarks=true]{hyperref}
\usepackage{amsmath}

\usepackage{caption}
\usepackage{threeparttable}

\usepackage{graphicx}

\begin{document}

\title{Active Surface-Driven Reconfigurable Gripper: \\Robust Grasping and Sequential Manipulation of Thin Objects}

\author{\authorblockN{Ziyi Zheng\textsuperscript{1,2},
Keqi Zhu\textsuperscript{1,2},
Hao Wu\textsuperscript{1,2}, 
Yanzhe Wang\textsuperscript{1,2}, 
Huixu Dong\textsuperscript{1,2,}\authorrefmark{2}}
\authorblockA{\textsuperscript{1}Grasp Lab, Zhejiang University
\quad\textsuperscript{2}Torch Kernel Co., Ltd.\\
\quad\authorrefmark{2}Corresponding author, e-mail: huixudong@zju.edu.cn}
}



%

\maketitle

\begin{abstract}
Robotic grippers face substantial challenges in grasping and manipulating thin objects. Most existing grippers rely on highly precise approach and grasp motions, which limits robustness and reduces applicability. This paper explores thin-object grasping using books as a representative example. Here, we propose a novel solution that integrates an active surface with underactuated compliance to achieve stable grasping of thin objects without complex control. First, an underactuated gripper with an active surface is designed. The active-surface thumb performs in-hand repositioning of the target book without requiring adjustments of the robot arm or the other fingers, while the underactuated fingers establish compliant contact conditions with the environment, and the reconfigurable structure enables reliable grasping of books under different configurations. Second, we establish a kinematic model of the gripper, and determine the initial grasp postures for two representative scenarios (books lying flat on a desktop and books vertically packed in a shelf). Third, by analyzing the physical model of a book lying on a table and its interaction with the gripper and the environment, we systematically optimize the structural parameters and grasping strategy. Finally, extensive experiments validate the effectiveness of the proposed gripper and strategy. The results demonstrate strong robustness and adaptability when grasping thin objects placed flat (including books, paper, fabric, plastic film, and mouse pad), as well as a high success rate when grasping vertically packed books. Moreover, the proposed gripper can reliably complete long sequential “grasp-place” tasks.
\end{abstract}

\IEEEpeerreviewmaketitle

\section{Introduction}
Robotic gripper plays a significant role in the interaction between robots and the environment, such as fruit picking \cite{11237059}, underwater operations \cite{gong2021soft}, and clearing dining plates \cite{shin2024bts}. Although robotic grippers have made remarkable progress in grasping and manipulating objects \cite{huang2025dih}, \cite{si2024deltahands}, \cite{11241047}, many scenarios in everyday life still require robotic grippers to be capable of handling thin deformable objects, such as the automated management of books in libraries, cafes, and offices. Therefore, developing a gripper capable of grasping thin and deformable objects across multiple scenarios is particularly important.

Typically, the thickness of thin deformable objects is much smaller than their length ($h/l < 1/10$) \cite{pilkey2002analysis}, and they have the characteristic of being easily deformable under force. However, grasping thin deformable objects presents significant challenges due to their thin and highly deformable nature \cite{9800969}, \cite{zheng2022autonomous}, \cite{guo2025enabling}. Specifically, grippers need to maintain contact with the surface during grasping, while the characteristics of thin deformable objects make the grasping motion difficult to predict. In this study, we select books as the research object. On one hand, books not only exhibit typical thin object deformability but also have the feature of separable layers, making the grasping process more challenging \cite{hasegawa2018detecting}. On the other hand, books in library scenarios exist in multiple states, such as lying flat on surfaces or vertically arranged on shelves. The robotic gripper needs to be versatile enough to handle books in these different scenarios.

\begin{figure}[t]
\vspace{-0mm}
\centerline{\includegraphics[width=1\columnwidth]{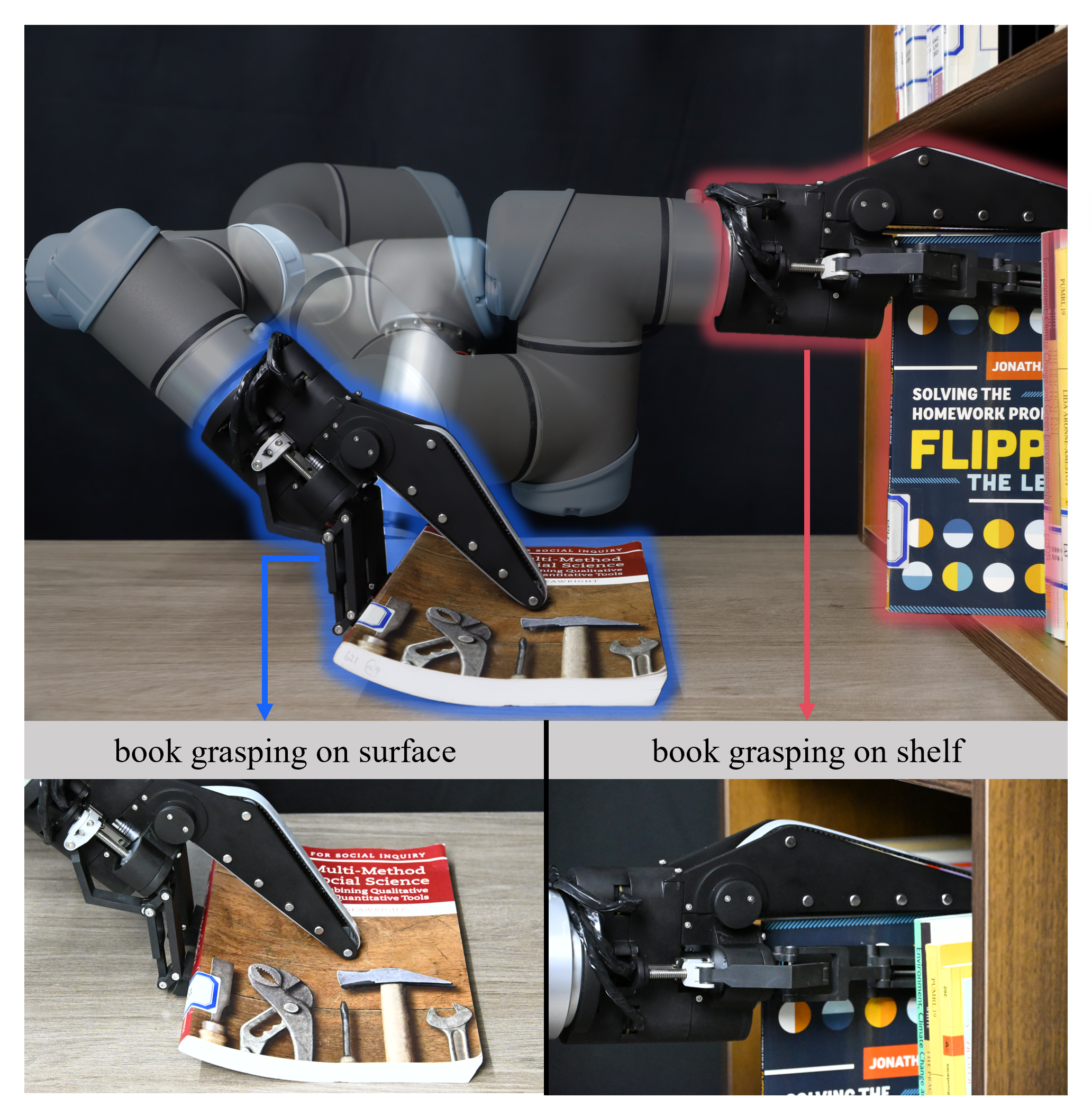}}
\vspace{-0mm}
\caption{\small Grasping demonstration. The gripper grasps books lying flat on the desktop and arranged on the bookshelf, respectively.}
\label{Fig:1}
\vspace{-0mm}
\end{figure}

Existing methods for grasping thin objects generally rely on precise finger gait control to maintain force closure during grasping \cite{zhang2022prying}, \cite{he2021scooping}. However, this approach greatly increases control complexity and depends on accurate mathematical models \cite{levesque2018model} or large-sample learning \cite{zhao2025learning}. Some studies have designed specialized grippers, such as using suction cups \cite{hasegawa2018detecting}, electro-adhesion \cite{sun2019new}, or external dexterity \cite{seino2025passive} to handle thin objects, but these methods impose significant requirements on the thickness and surface friction of the objects, lacking generality. Furthermore, existing methods typically focus on a single state of thin objects (lying flat), making it difficult to extend to multi-state grasping in daily scenarios. Although studies such as \cite{zhao2025learning} and \cite{jiang2025rotipbot} have explored grasping thin objects on inclined surfaces, their operational logic remains unchanged. For the two fundamentally different states of books, it is necessary to develop a simple and general gripper.

To address the challenges of grasping thin and deformable objects, in this study, we propose a reconfigurable gripper with an active surface and a grasping strategy that can simultaneously handle books in two states commonly found in libraries, as shown in Fig. \ref{Fig:1}. Specifically, the proposed gripper consists of three fingers, with the thumb equipped with a belt as the active surface for repositioning books during the grasping process. The other two fingers have coupled opening and restructuring movements to achieve the design goal of handling books in two different states. A spring connects the fingertip to the second phalanx, giving the gripper passive compliance with the environment. Furthermore, we analyze the kinematics of the gripper and propose a grasping strategy based on it. The grasping action of the proposed gripper mainly involves three steps: determining the initial configuration, repositioning the book, and parallel grasping. Unlike traditional finger gait strategies, the repositioning step uses the continuously rolling active surface to provide a stable contact point for the book, enhancing robustness and reducing control complexity. Additionally, we optimize the gripper's design parameters and grasping strategy based on the beam model. Finally, systematic experiments validate the rationality of the proposed gripper design and strategy. The experimental results demonstrate that the gripper exhibits good grasping stability and versatility, providing a universal solution for book retrieval in libraries. Furthermore, our strategy can be extended to the grasping of various thin objects, offering a new and straightforward control method for grasping thin objects.

The main contributions of this paper include:

1) We propose a novel gripper that integrates an active surface structure, reconfigurable mechanisms, and underactuated joints, achieving good generality with simple control.

2) We develop an active surface-based grasping strategy. The gripper only needs to determine the initial configuration and uses the active surface to reposition thin deformable objects, avoiding complex finger gait control during grasping.

3) Based on beam theory, we analyze the contact situation during the grasping process, further optimizing the structural parameters and grasping strategy.

4) We conduct systematic experiments to evaluate the designed gripper, demonstrating its ability to stably grasp both separated and single-layer thin objects, as well as complete long sequences of “grasp-place” tasks.

The rest of this paper is structured as follows. Section II reviews related work; Section III introduces the gripper structure; Section IV conducts the kinematic analysis; Section V proposes the grasping strategy and parameter optimization; Section VI presents the experiments. Section VII concludes the paper.

\section{Related Works}

In recent years, the manipulation of thin deformable objects has become a rapidly growing research area \cite{zheng2022autonomous}, \cite{lee2024gog}, \cite{wang2025sa}. However, grasping thin deformable objects such as books still faces numerous challenges. Some studies utilize pre-grasp actions to move the object to the edge of a surface before completing the grasp \cite{hang2019pre}, \cite{ghazaei2020quasi}, which requires the prediction of object trajectories and significant mechanical arm motion space. Scooping is another effective strategy for grasping thin objects \cite{he2021scooping}, \cite{babin2018picking}, where the fingers are inserted into the gap between the thin object and the surface to complete the grasp \cite{babin2019stable}, \cite{do2024densetact}. However, the contact points between the object and fingers continuously change during the grasping process, making it more difficult to maintain force closure, typically requiring coordinated motion of the robotic arm and fingers. In \cite{levesque2018model}, a model-based scooping strategy was proposed to achieve force-closure grasping of thin objects. In \cite{wang2021gelsight}, the GelSight Wedge sensor was developed to obtain contact point information. Similarly, prying methods involve placing two fingers on either side of the thin object, lifting the object before grasping it \cite{zhang2022prying}, \cite{tong2020picking}. Like scooping, prying also requires precise finger or robotic arm motion control.

Soft grippers, can better handle uncertainties during grasping, due to their passive compliance \cite{teeple2022multi}, \cite{tran2025hybrid}, yet prediction of finger motion remains necessary. In \cite{tran2025hybrid}, the rotation angle and stiffness of the soft gripper were experimentally measured. In \cite{jiang2019dynamic}, an energy-based method was used to predict the contact between flexible fingers and objects. In \cite{zhao2025learning}, a two-finger soft gripper equipped with force/torque sensors was trained using reinforcement learning to perform grasping. Although these soft grippers demonstrate good robustness when grasping single thin sheets, their strategies typically involve contacting the upper surface of thin objects, making it challenging to grasp multilayer separable thin soft objects such as books.

Additionally, some specialized grippers have been developed to handle thin deformable objects using suction cups \cite{hasegawa2018detecting}, \cite{chen2025versatile}, electro-adhesion \cite{sun2019new}, or external dexterity \cite{seino2025passive}, avoiding the need for finger gait control during grasping. However, these grippers impose stringent requirements on the properties of the objects. Suction cups and electro-adhesion demand absolutely smooth surfaces to generate sufficient suction or adhesion force \cite{sun2019new}, \cite{chen2025versatile}. Meanwhile, grippers relying on external dexterity struggle to handle thin objects with high bending stiffness \cite{seino2025passive}.

\begin{figure*}[t]
\vspace{-0mm}
\centerline{\includegraphics[width=2\columnwidth]{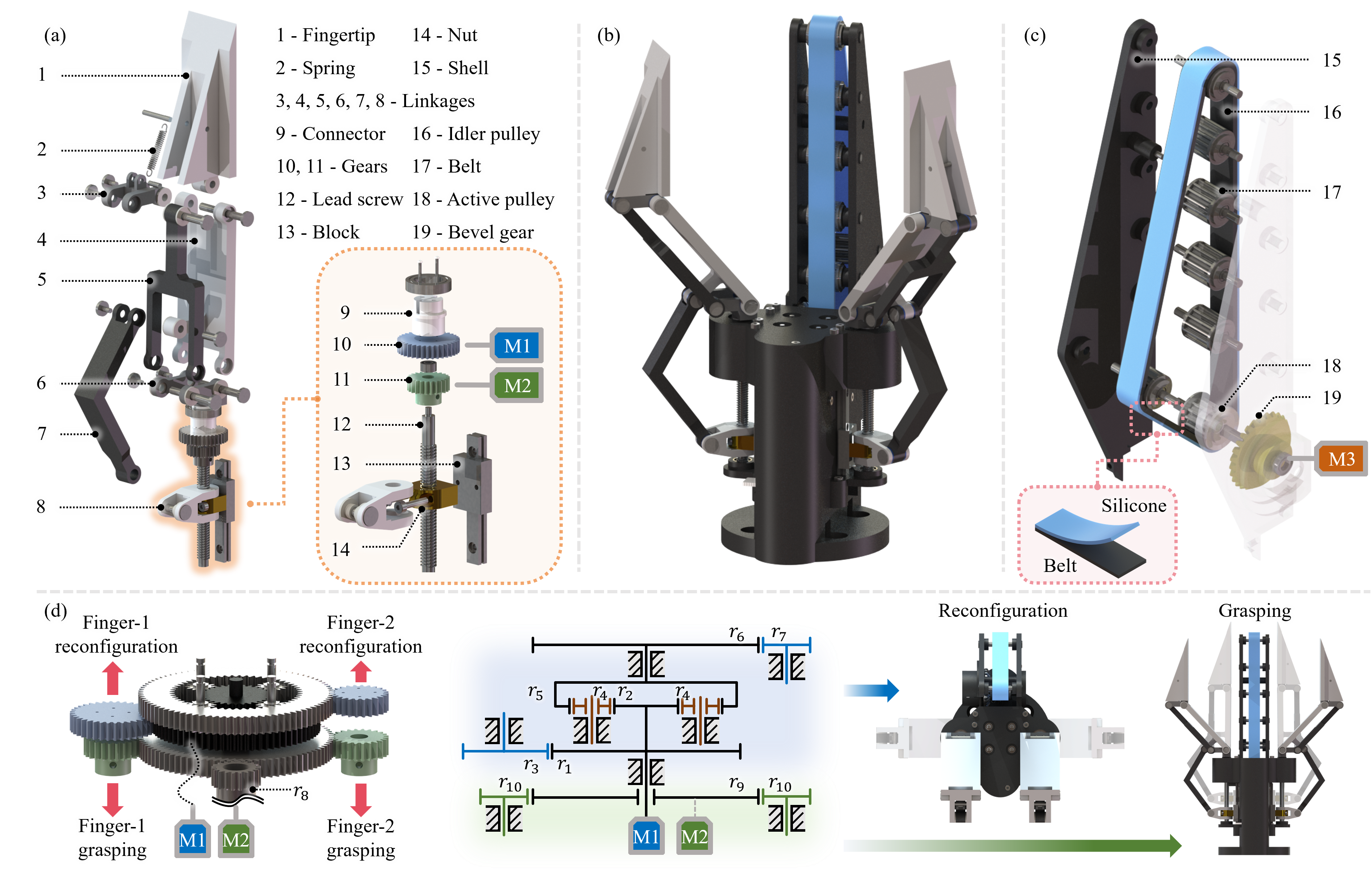}}
\vspace{-0mm}
\caption{\small The structural design of the proposed gripper. (a) Explosion diagram of underactuated finger. (b) Overall configuration of the proposed gripper. (c) Explosion diagram of the active surface thumb. (d) Transmission structure and its implementation of reconfiguration and grasping motions.}
\label{Fig:2}
\vspace{-0mm}
\end{figure*}

Some studies have equipped grippers with active surfaces to perform grasping and manipulation tasks without depending on finger gait control \cite{he2025grasping}, \cite{xie2023hand}. These active surfaces are typically composed of belts \cite{10251543}, \cite{wang2025reconfigurable} or rollers \cite{yuan2024design}, \cite{yuan2020design}. In the field of grasping thin objects, active surface grippers simplify control strategies due to their straightforward operation. In \cite{yamazaki2021versatile}, a gripper used short bristles on rollers to pick up fabric through a single rolling motion, but this method is not suitable for handling smooth-surfaced objects. To broaden applicability, the grippers in \cite{jiang2025rotipbot}, \cite{unde2024design} were equipped with rollers on two fingers that rotate in opposite directions to lift protrusions on thin and deformable objects before grasping. However, this method is limited to handling thin objects with small bending stiffness due to significant object deformation. In \cite{ko2020tendon}, a 1-DOF gripper with a belt-based active surface was developed to handle thin objects with higher bending stiffness, such as books. Nevertheless, the coupling of active surface and finger movements necessitates adjustments to the robotic arm during the grasping process. Therefore, developing a simple operational method for handling a wide variety of thin and deformable objects remains challenging.

\section{Gripper Design}
\subsection{Task Analysis}
The gripper proposed in this study is designed with the objective of grasping books in different states. In daily life scenarios such as libraries and cafes, two typical book states are frequently encountered: books lying flat on a surface and books arranged vertically on a bookshelf. Books lying flat can be regarded as thin deformable objects with separable layers, exhibiting significant variations in thickness and surface friction among different books. For this case, approaches that manipulate only a single surface can easily lead to interlayer separation \cite{yamazaki2021versatile}, \cite{unde2024design}. Therefore, a grasping strategy utilizing an active surface is developed. Specifically, friction between the active surface and the book is exploited to move the book toward the inner side of the gripper. During this process, continuous rolling of the active surface provides a stable and continuously changing contact point on the book, which remains fixed relative to the gripper coordinate frame. This characteristic effectively avoids the need for complex finger motion control during book repositioning. After the book is repositioned, the fingers close to complete the grasp. In another scenario, for books vertically arranged on a bookshelf, the active surface is also employed to reposition the target book. In this case, the book can be regarded as a rigid body under forces applied along its width direction. However, the target book may be tightly wedged between adjacent books, which could unintentionally pull neighboring books out. To address this, a separation strategy for adjacent books is proposed. Specifically, the two fingers are inserted into the gaps between the target book and surrounding books, and the active surface contacts the side of the target book (the side facing upward) to perform repositioning. Subsequently, the gripper retracts while firmly grasping the book using the high-friction active surface.

To realize these grasping tasks, it is necessary to design a gripper that concurrently satisfies both types of scenarios. This study proposes a reconfigurable three-finger gripper with an active surface, as shown in Fig. \ref{Fig:2}. Unlike traditional active surface grippers that are equipped with active surfaces on all fingers, the proposed gripper features an active surface only on the thumb, while the tips of the other two fingers adopt a wedge-shaped structure. This design aims to facilitate the smooth repositioning of thin deformable objects like books. Additionally, the reconfigurable feature allows two fingers to rotate symmetrically relative to the base, enabling the gripper to switch between grasping and separation modes. To simplify control and reduce the number of actuators, the thumb is fixed, while the opening/closing motions of the two fingers are coupled, as is the reconfiguration motion. Therefore, the entire gripper functionality is realized with only three motors: motor 1 controls the finger reconfiguration, motor 2 controls the opening/closing motion, and motor 3 controls the rotation of the active surface.

\subsection{Mechanical Structure Design}
\subsubsection{Underactuated Fingers}
For handling objects with regular shapes like books, parallel grasping capability of the gripper is essential to increase the contact area between the gripper and the book for stable grasping. Fig. \ref{Fig:2}(a) shows the basic structure of the finger, where the upper part of the finger includes a parallel four-bar linkage structure composed of links 3, 4, 5, and 6 to achieve parallel grasping. Link 6 is connected to a gear (blue) via a connector and is driven by motor 1 to achieve finger reconfiguration. Since smooth repositioning is necessary for books lying flat on a surface, and the fingers need to be inserted into the gaps between vertically arranged books, the fingertip is designed to be wedge-shaped. The fingertip is connected to link 3 by a spring, forming an underactuated joint. When the back of the fingertip experiences external force, the finger can passively bend inwards. This feature allows the gripper to passively bend the fingertips based on the contact situation when grasping books on a flat surface, eliminating the need for additional active degrees of freedom to control the finger structure. In its natural state or during grasping, the fingertip is tightened by the spring to a posture perpendicular to the palm, due to the limiting structure between the fingertip and link 3. The opening and closing motion of the finger is controlled by motor 2, where the input gear (green) is fixed to the lead screw. The rotational motion is converted to linear motion using the lead screw and nut, and is transmitted to the parallel four-bar linkage mechanism through links 8, 7, and 4. Notably, link 8 is designed as a sleeve structure, with its vertical movement restricted by the nut while allowing rotational motion. This enables the decoupling of finger reconfiguration and opening/closing motion.
\subsubsection{Active Surface Thumb}
Various methods can be employed to achieve continuous rotation of the gripper’s active surface, such as roller \cite{yuan2024design}, cylinder \cite{8206013}, and belt \cite{10251543}. In this study, considering grasping stability, a belt is selected as the active surface to provide a larger contact area after grasping books, as shown in Fig. \ref{Fig:2}(c). Here, motor 3 drives the belt motion through bevel gear transmission and active pulley to achieve the repositioning of the book. To enhance friction, the belt surface is coated with a layer of silicone. Additionally, six idler pulleys are installed on the inner side of the belt, with five of them densely arranged on the contact surface to ensure a larger contact area between the book and the thumb after grasping, providing a more stable grasping effect.

\begin{figure}[t]
\vspace{-0mm}
\centerline{\includegraphics[width=1\columnwidth]{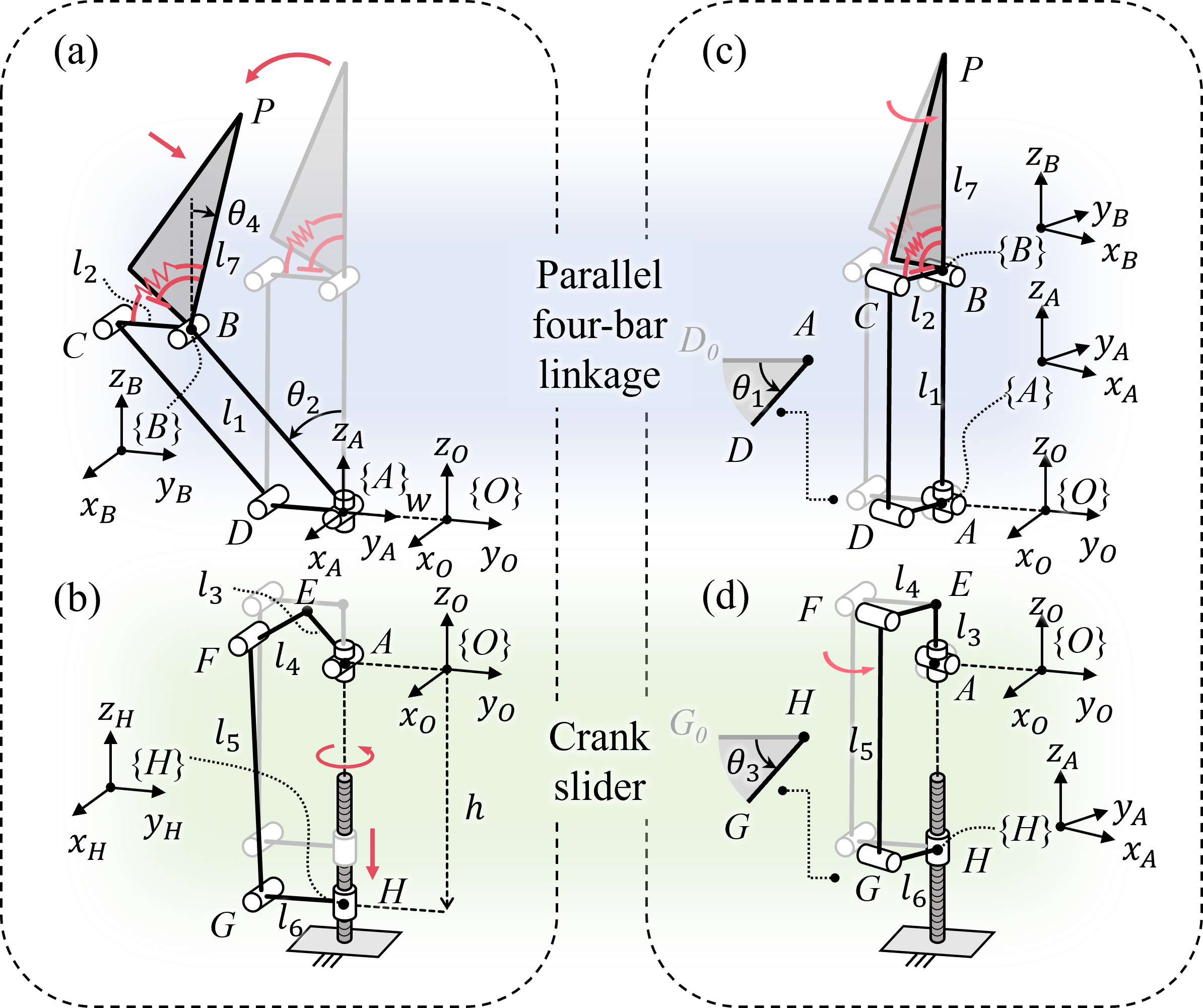}}
\vspace{-0mm}
\caption{\small Finger kinematics analysis. (a) Grasping motion with a parallel four-bar linkage. (b) Grasping motion with a crank-slider mechanism. (c) Reconfiguring motion with a parallel four-bar linkage. (d) Reconfiguring motion with a crank-slider mechanism.}
\label{Fig:3}
\vspace{-0mm}
\end{figure}

\begin{table}[b]
\centering
\renewcommand{\arraystretch}{1.5}
\caption{Gear Parameters}
\label{tab:table1}
\begin{tabular}{ccccccccccc}
\hline
Parameters & r1 & r2 & r3 & r4 & r5 & r6 & r7 & r8 & r9 & r10 \\ \hline
Values(mm) & 18 & 8  & 9  & 3  & 14 & 21 & 6  & 5  & 21 & 6   \\ \hline
\end{tabular}
\end{table}

\subsubsection{Transmission Mechanism}
To achieve decoupled motion between finger opening/closing and reconfiguration, the gripper's transmission mechanism is divided into two independent layers. The upper layer is responsible for changing the finger configuration, while the lower layer executes the grasping action. These two gear sets are separately driven by two motors for independent transmission, as illustrated in Fig. \ref{Fig:2}(d). The upper gear set forms a planetary gear system. Unlike a traditional planetary gear system, the circular rotation of the four planet gears is restricted. Additionally, an external gear meshes with the gear (blue) on finger 2. This design enables the two fingers to rotate symmetrically by the same angle. Assuming the angular velocity input to the planetary gear system is $\omega_{in}$, the angular velocity of gear 3 can be expressed as follows:
\begin{equation}
\label{eqn:1}
\omega_{3}=\omega_{i n} \frac{r_{1}}{r_{3}}
\end{equation}

The angular velocity of gear 7 is represented as:
\begin{equation}
\label{eqn:2}
\omega_{7}=\omega_{i n} \frac{r_{2} r_{6}}{r_{5} r_{7}}
\end{equation}

To ensure that both fingers rotate symmetrically by the same angle, the angular velocities of gear 3 and gear 7 need to be the same, and the meshing conditions of the planetary gear system must be satisfied, i.e.,
\begin{equation}
\label{eqn:3}
\left\{\begin{array}{l}
\omega_{3}=\omega_{7} \\
r_{2}+2 r_{4}=r_{5}
\end{array}\right.
\end{equation}

Furthermore, since the planar crank-slider mechanism requires gear 3 and gear 10 (left) to be coaxial, as well as gear 7 and gear 10 (right), the two-layer gear transmission mechanisms need to satisfy the following conditions:
\begin{equation}
\label{eqn:4}
r_{1}+r_{3}=r_{9}+r_{10}=r_{6}+r_{7}
\end{equation}

Based on the above constraints, a set of gear parameters is obtained, as shown in the Table \ref{tab:table1}.

\section{Kinematics Analysis of The Gripper}
As described in Section III, the proposed gripper features two decoupled motions for grasping and reconfiguring. The grasping motion is achieved through the coordinated movement of the crank-slider mechanism and the parallel four-bar linkage, while the reconfiguring motion is directly driven by the motor through a planetary gear set.

The gripper coordinate system $\{O\}$ has its origin at the midpoint of the line connecting the bases of the two fingers, the proximal joint coordinate system $\{A\}$ has its origin at the midpoint of the rotation joint of the proximal joint, and the distal joint coordinate system $\{B\}$ has its origin at the midpoint of the rotation joint of the distal joint. The slider coordinate system $\{H\}$ has its origin at the center of the nut fixed on the slider, as shown in Fig. \ref{Fig:3}. The initial configuration is defined by the relative positions of the two fingers and the proximal joint being perpendicular to the $yz$ plane.

The transformation matrix from the gripper coordinate system $\{O\}$ to the proximal joint coordinate system $\{A\}$ can be obtained as ${ }^{O} \boldsymbol{T}_{A}$, and the transformation matrix from the proximal phalanx coordinate frame to the endpoint $F$ is denoted as ${ }^{A} \boldsymbol{T}_{F}$. Therefore, the transformation matrix from the gripper coordinate frame to endpoint $F$ is given by ${ }^{O} \boldsymbol{T}_{F}={ }^{O} \boldsymbol{T}_{A}{ }^{A} \boldsymbol{T}_{F}$. From which the coordinates of endpoint $F$ in the gripper coordinate frame can be obtained as ${ }^{O} \boldsymbol{p}_{F}$. On the other hand, the transformation matrix from the gripper coordinate frame to the slider coordinate frame is ${ }^{O} \boldsymbol{T}_{H}$, and the transformation matrix from the slider coordinate frame to endpoint $G$ is ${ }^{H} \boldsymbol{T}_{G}$. Thus, the transformation matrix from the gripper coordinate frame to endpoint $G$ is ${ }^{O} \boldsymbol{T}_{G}={ }^{O} \boldsymbol{T}_{H}{ }^{H} \boldsymbol{T}_{G}$, and the coordinates of endpoint $G$ in the gripper coordinate frame, ${ }^{O} \boldsymbol{p}_{G}$, can be calculated. According to the constraint relations of the planar crank-slider mechanism, the finger rotation angle $\theta_{2}$ can be expressed in terms of the slider displacement $h$. Focusing on the upper part of the finger, the transformation matrix from the proximal phalanx coordinate frame to the distal phalanx coordinate frame is ${ }^{A} \boldsymbol{T}_{B}$, and from the distal phalanx coordinate frame $\{B\}$ to endpoint $P$ is ${ }^{B} \boldsymbol{T}_{P}$. Therefore, the coordinate transformation from the gripper coordinate frame $\{O\}$ to endpoint $P$ is ${ }^{O} \boldsymbol{T}_{P}={ }^{O} \boldsymbol{T}_{A}{}^{A} \boldsymbol{T}_{B}{}^{B} \boldsymbol{T}_{P}$. From which the coordinates of endpoint $P$ in the gripper coordinate frame, ${ }^{O} \boldsymbol{p}_{P}$, can be obtained. Similarly, the coordinates of endpoint $Q$ in the gripper coordinate frame, ${ }^{O} \boldsymbol{p}_{Q}$, can be derived, i.e.,
\begin{equation}
\label{eqn:20}
{ }^{O} \boldsymbol{p}_{P}=\left[\begin{array}{c}
l_{1} \sin \theta_{1} \sin \theta_{2}-l_{7} \sin \theta_{1} \sin \theta_{4} \\
l_{7} \cos \theta_{1} \sin \theta_{4}-l_{1} \cos \theta_{1} \sin \theta_{2}-w \\
l_{7} \cos \theta_{4}+l_{1} \cos \theta_{2}
\end{array}\right]
\end{equation}

\begin{equation}
\label{eqn:21}
{ }^{O} \boldsymbol{p}_{Q}=\left[\begin{array}{c}
l_{1} \sin \theta_{1} \sin \theta_{2}-l_{7} \sin \theta_{1} \sin \theta_{4} \\
-l_{7} \cos \theta_{1} \sin \theta_{4}+l_{1} \cos \theta_{1} \sin \theta_{2}+w \\
l_{7} \cos \theta_{4}+l_{1} \cos \theta_{2}
\end{array}\right]
\end{equation}

Please see Appendix A for the detailed calculation process.

\section{Grasping Strategy and Parameter Optimization}
\subsection{Grasping Strategy}

\begin{figure}[t]
\vspace{-0mm}
\centerline{\includegraphics[width=1\columnwidth]{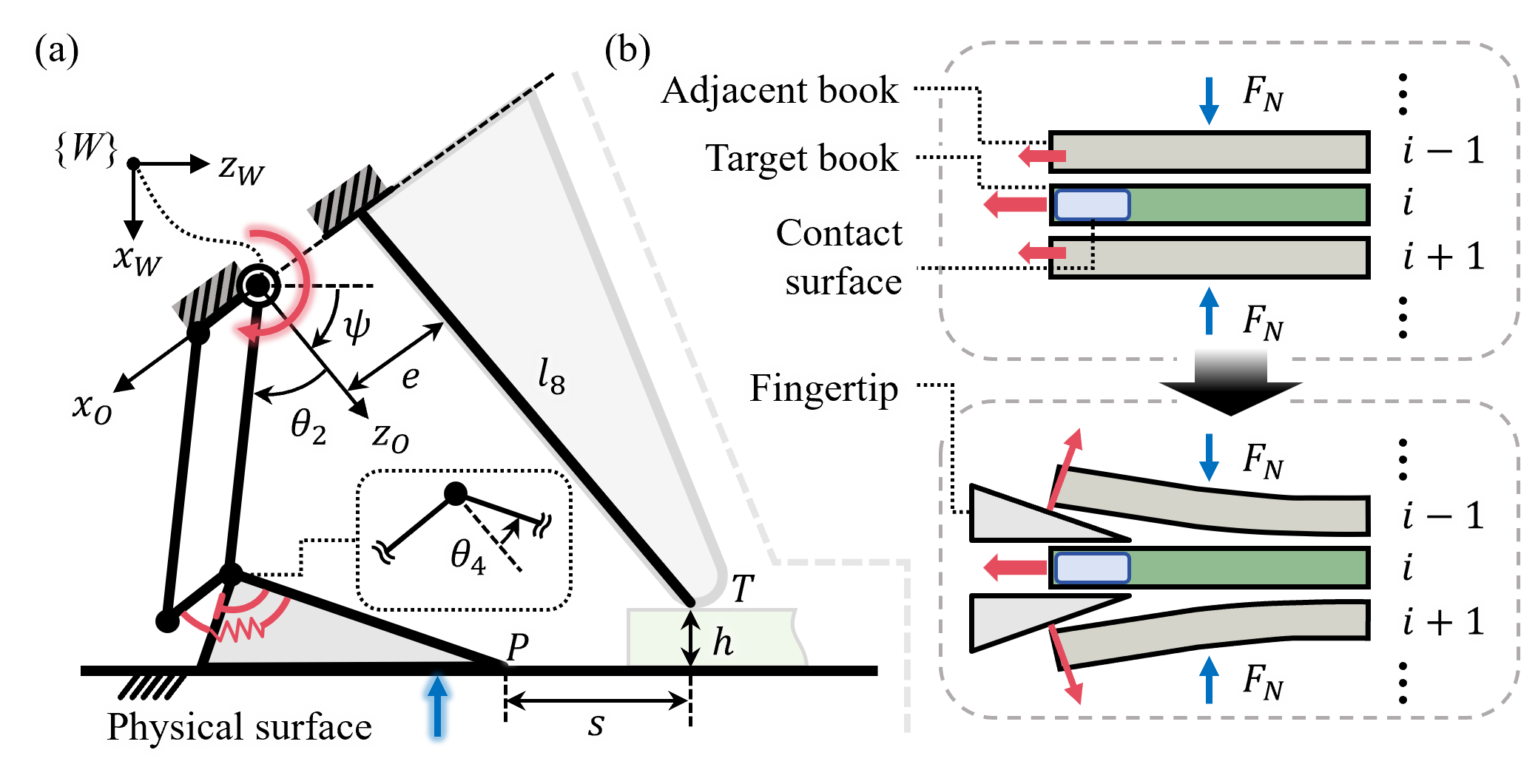}}
\vspace{-0mm}
\caption{\small Grasping strategies. (a) Initial configuration for grasping books lying on a surface. (b) Comparison of two grasping strategies for books on the bookshelf (top view).}
\label{Fig:4}
\vspace{-0mm}
\end{figure}

\subsubsection{Grasping Strategy for Flat Books}
Thanks to the design of the active surface, the proposed gripper only needs to determine the initial configuration without requiring adjustments to the gripper’s orientation or finger movements during the manipulation of the book. Fig. \ref{Fig:4}(a) shows the initial configuration of the gripper when grasping a book with a thickness of  $h$. At this moment, the thumb presses against the upper surface of the book, while the two underactuated fingers passively adjust their rotation angles to conform to the surface. In this configuration, the surfaces of the fingers are parallel. Taking finger $P$ as the analysis subject, the expression of eqn. (\ref{eqn:20}) can be derived in the $xz$ plane as follows:
\begin{equation}
\label{eqn:22}
{ }^{O} \boldsymbol{p}_{P, x z}=\left[\begin{array}{l}
l_{1} \sin \theta_{2}-l_{7} \sin \theta_{4} \\
l_{1} \cos \theta_{2}+l_{7} \cos \theta_{4}
\end{array}\right]
\end{equation}

Due to the spring connection between the fingertip and the four-bar linkage mechanism, the angle $\theta_4$ is passively adjusted based on external environmental constraints. It can be expressed as $\theta_{4}=\psi-\alpha$, where $\psi$ is the rotation angle of the gripper, and $\alpha$ is the angle of the fingertip. Taking the rotation center of the gripper, i.e., the base of the finger, as the origin of the world coordinate system $\{W\}$, the coordinates of the fingertip $P$ in the world coordinate system can be obtained through a rotation matrix ${ }^{W} \boldsymbol{R}_{O}$:
\begin{equation}
\label{eqn:23}
\begin{array}{l}
{ }^{W} \boldsymbol{p}_{P, x z}={ }^{W} \boldsymbol{R}_{O}^{ }{ }^{O} \boldsymbol{p}_{P, x z}=\left[\begin{array}{l}
x_{P} \\
z_{P}
\end{array}\right] \\
=\left[\begin{array}{l}
l_{1} \sin \left(\theta_{2}+\psi\right)+l_{7} \sin \left(\psi-\theta_{4}\right) \\
l_{1} \cos \left(\theta_{2}+\psi\right)+l_{7} \cos \left(\theta_{4}-\psi\right)
\end{array}\right]
\end{array}
\end{equation}

In the $xz$ plane, the coordinates of the thumb tip $T$ in the world coordinate system can be expressed as follows:
\begin{equation}
\label{eqn:24}
\begin{array}{l}
{ }^{W} \boldsymbol{p}_{T, x z}=\left[\begin{array}{c}
x_{T} \\
z_{T}
\end{array}\right]=\left[\begin{array}{c}
-e \cos \psi+l_{8} \sin \psi \\
e \sin \psi+l_{8} \cos \psi
\end{array}\right]
\end{array}
\end{equation}
where, $e$ represents the distance between the base of the finger and the thumb, and $l_8$ denotes the length of the thumb. By combining eqn. (\ref{eqn:23}) and eqn. (\ref{eqn:24}), the vertical distance $h$ (i.e., the thickness of the book) and the horizontal distance $s$ between the fingertip and the thumb tip can be obtained as follows:
\begin{equation}
\label{eqn:25}
\begin{array}{l}
\left\{\begin{array}{l}
g_{x}\left(\theta_{2}, \psi\right)=x_{P}-x_{T}=h \\
g_{z}\left(\theta_{2}, \psi\right)=z_{T}-z_{P}=s
\end{array}\right.
\end{array}
\end{equation}

Therefore, once $h$ and $s$ are determined, the initial configuration of the gripper (i.e., $\theta_{2}$ and $\psi$) can be obtained. Subsequently, the repositioning and grasping manipulations can be performed.

\subsubsection{Grasping Strategy for Vertically Arranged Books}
Unlike traditional methods that rely on finger gait to grasp books on shelves, the proposed gripper can also utilize the active surface to grasp books on the bookshelf. In this scenario, the gripper is in separation mode, where the two fingers should align with the gap between the target book and the surrounding books, inserting the fingertips into the gap. The wedge-shaped fingertips are then used to separate the target book from its neighbors. The purpose of this step is to reposition the target book without disturbing the surrounding books, as shown in Fig. \ref{Fig:4}(b). Moreover, for books with smaller thickness, this separation ensures that the active surface on the thumb contacts only the target book. Similar to grasping books placed flat, the gripper only needs to determine the initial configuration. Let the thickness of the target book be $h$, and the initial configuration of the gripper is such that the two fingers are aligned. Therefore, in the $yz$ plane, eqn. (\ref{eqn:20}) and eqn. (\ref{eqn:21}) can be simplified as follows:
\begin{equation}
\label{eqn:26}
\begin{array}{l}
\begin{array}{c}
{ }^{O} \boldsymbol{p}_{P, y z}=\left[\begin{array}{c}
-l_{1} \sin \theta_{2}-w \\
l_{1}+l_{7}
\end{array}\right]
\end{array}
\end{array}
\end{equation}

\begin{equation}
\label{eqn:27}
\begin{array}{l}
\begin{array}{c}
{ }^{O} \boldsymbol{p}_{Q, y z}=\left[\begin{array}{c}
l_{1} \sin \theta_{2}+w \\
l_{1}+l_{7}
\end{array}\right]
\end{array}
\end{array}
\end{equation}

Therefore, the thickness of the target book corresponds to the distance between the two fingers, which can be expressed as:
\begin{equation}
\label{eqn:28}
\begin{array}{l}
2 l_{1} \sin \theta_{2}+2 w=h
\end{array}
\end{equation}

Considering the need to translate the fingers into the gap between the target book and the adjacent books, the gripper’s rotation angle $\psi$ is set to 0, thereby defining the initial configuration of the gripper. After the gripper completes the insertion action, the active surface is then used to reposition the book. The fingers are then reconfigured, switching to the grasping mode. In this configuration, the high-friction active surface contacts the book, allowing for a more stable grasp.

\subsection{Optimization of Structural Parameters and Grasping Strategies}
\begin{figure}[t]
\vspace{-0mm}
\centerline{\includegraphics[width=1\columnwidth]{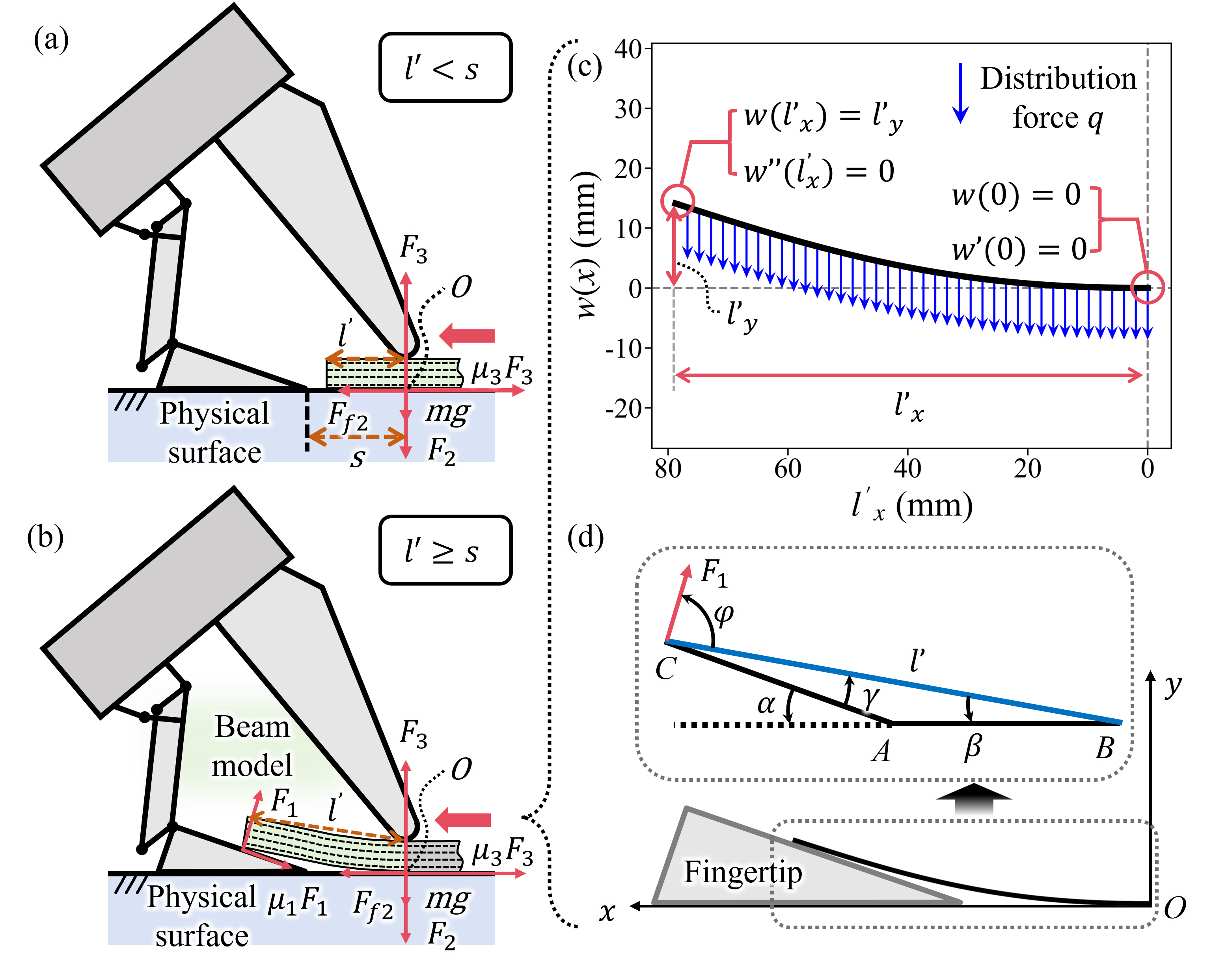}}
\vspace{-0mm}
\caption{\small Contact force analysis during repositioning operation. (a) Stage before the book contacts the fingers. (b) Stage after the book contacts the fingers. (c) Forces on the book and boundary conditions. (d) Approximation of line segment length.}
\label{Fig:5}
\vspace{-0mm}
\end{figure}

To successfully grasp a book placed flat, the active surface of the thumb must not experience relative slipping during the repositioning process of the book, that is, $F_{f 2} \leq \mu_{2} F_{2}$, here, $F_{f 2}$ denotes the friction force exerted by the thumb on the book, $\mu_{2}$ is the friction coefficient between the thumb and the book, and $F_{2}$ is the normal force applied by the thumb on the book (the explanations of all control and object parameters can be found in Table I in Appendix B). Meanwhile, the system must satisfy force equilibrium conditions. The grasping process can be regarded as quasi-static. For a book with length $l$, width $b$, thickness $h$, and weight $mg$, at any moment during the grasping process, $l^{\prime}$ denote the length of the book to the left of the thumb contact point. When $l^{\prime}<s$, the book translates on the table without contacting the fingers, as shown in Fig. \ref{Fig:5}(a), and the following equilibrium equations hold:
\begin{equation}
\label{eqn:29}
\begin{array}{l}
F_{f 2}=\mu_{3} F_{3}=\mu_{3} m g+\mu_{3} F_{2}
\end{array}
\end{equation}
where, $\mu_{3}$ is the friction coefficient between the book and the table, and $F_{3}$ is the normal force exerted by the table on the book. When $l^{\prime} \geq s$, the book comes into contact with the fingers and is subjected to the finger’s normal force $F_{1}$ and friction force, as shown in Fig. \ref{Fig:5}(b). Based on force equilibrium, we can get:
\begin{equation}
\label{eqn:30}
\begin{array}{l}
\begin{array}{l}
F_{f 2}=\mu_{3} m g+\mu_{3} F_{2} +KF_{1}
\end{array}
\end{array}
\end{equation}
where $K=\sin \alpha+\mu_{1} \cos \alpha-\mu_{3} \cos \alpha+\mu_{1} \mu_{3} \sin \alpha$, $u_{1}$ is the friction coefficient between the book and the fingers, and $\alpha$ is the angle of fingertip. Therefore, a larger friction force occurs during the stage when the book is in contact with the fingers. In this stage, the portion of the book to the left of the thumb maintains moment equilibrium. Approximating the line segment $BC$ as $l^{\prime}$, which connects the book-finger contact point and the coordinate origin, we obtain:
\begin{equation}
\label{eqn:31}
\begin{array}{l}
F_{1}=\left(M_{0}+m^{\prime} g \cos \beta \cdot l^{\prime} / 2\right) /\left(\sin \varphi \cdot l^{\prime}-\mu_{1} \sin \alpha \cdot s\right)
\end{array}
\end{equation}
where, $M_{0}$ is the bending moment relative to the origin. $m^{\prime} g$ is the weight of the portion of the book to the left of the thumb contact point, i.e., $m^{\prime} g=m g\left(l^{\prime} / l\right)$, $\beta$ is the angle between segment $BC$ and the $x$-axis, and $\varphi$ is the angle between $F_{1}$ and $BC$. As illustrated in Fig. \ref{Fig:5}(d), it can be derived:
\begin{equation}
\label{eqn:32}
\begin{array}{l}
\cos \beta=\cos \alpha \cdot \cos \gamma+\sin \alpha \cdot \sin \gamma, \\
\sin \varphi=\sin \left(\frac{\pi}{2}-\gamma\right)=\cos \gamma
\end{array}
\end{equation}
where the angle $\gamma$ is the angle between $BC$ and $AC$, with $\sin \gamma=s \cdot \sin \alpha / l^{\prime}$, $\quad \cos \gamma=\sqrt{1-\left(s \cdot \sin \alpha / l^{\prime}\right)^{2}}$.

Considering that the thin objects studied in this paper, such as books, belong to a typical slender structure ($h/l < 1/10$), and that the deformation during the repositioning process is relatively small, the Euler-Bernoulli beam model \cite{pilkey2002analysis} is used as a first-order approximation. The bending moment $M(x)$ of the book at position $x$ can be expressed by the curvature $\omega^{\prime \prime}(x)$, i.e.,
\begin{equation}
\label{eqn:33}
\begin{array}{l}
M(x)=E I_{z} \omega^{\prime \prime}(x)
\end{array}
\end{equation}
where $E$ is the Young’s modulus, and $I_{z}$ is the moment of inertia of the book, which can be expressed as:
\begin{equation}
\label{eqn:34}
I_{z}=\frac{b h_{\text {paper }}^{3}}{12} \cdot \frac{h}{h_{\text {paper }}}
\end{equation}
where $h_{\text {paper }}$ represents the thickness of a single sheet of paper. Since the mass distribution of the book is uniform, the distributed load can be expressed as:
\begin{equation}
\label{eqn:35}
E I_{z} \omega^{\prime \prime \prime \prime}(x)=-q=-\frac{m g}{l}
\end{equation}

By performing four successive integrations, the deflection curve equation with four parameters $A$, $B$, $C$, and $D$ can be obtained:
\begin{equation}
\label{eqn:36}
\omega(x)=-\frac{q}{24 E I_{z}} x^{4}+A x^{3}+B x^{2}+C x+D
\end{equation}

During the quasi-static process, the right end of the book ($x=0$) can be considered fixed, and the left end ($x=l_{x}^{\prime}$) is free. Therefore, the following four boundary conditions can be established: the deflection at the right end is 0 ($\omega(0)=0$); the slope at the right end is 0 ($\omega^{\prime}(0)=0$); the deflection at the left end is $l_{y}^{\prime}$ ($\omega(l_{x}^{\prime})=l_{y}^{\prime}$); and the curvature at the left end is 0 ($\omega^{\prime\prime}(l_{x}^{\prime})=0$), as shown in Fig. \ref{Fig:5}(c). Combining these four boundary conditions with eqn. (\ref{eqn:36}), we obtain:
\begin{equation}
\label{eqn:37}
\left\{\begin{array}{l}
A=\frac{5 q}{48 E I_{z}} l_{x}^{\prime}-\frac{l_{y}^{\prime}}{2 l_{x}^{\prime}{ }^{3}} \\
B=-\frac{q}{16 E I_{z}} l_{x}^{\prime}{ }^{2}+\frac{3 l_{y}^{\prime}}{2 l_{x}^{\prime}{ }^{2}} \\
C=D=0
\end{array}\right.
\end{equation}

Therefore, the bending moment at the right end can be obtained as:
\begin{equation}
\label{eqn:38}
M_{0}=E I_{z} \omega^{\prime \prime}(0)=E I_{z}\left(-\frac{m g}{8 E I_{z} l} l_{x}^{\prime}{ }^{2}+\frac{3 l_{y}^{\prime}}{l_{x}^{\prime}{ }^{2}}\right)
\end{equation}
where, $l_{x}^{\prime}{ }=l^{\prime}\cos\beta$, $l_{y}^{\prime}{ }=l^{\prime}\sin\beta$. By combining eqn. (\ref{eqn:30}), eqn. (\ref{eqn:31}), and eqn. (\ref{eqn:38}), the friction force $F_{f 2}$ exerted by the thumb on the book can be obtained. Therefore, the friction condition can be expressed as:
\begin{equation}
\label{eqn:39}
g\left(F_{1}\right) \leq g\left(F_{2}\right)
\end{equation}

During the grasping process, a smaller value of $F_2$ should be used to protect the gripper and the book. Therefore, $F_1$ should be reduced to ensure that eqn. (\ref{eqn:39}) holds, allowing the thumb to prevent relative slipping between the book and the thumb under lower pressure. Consequently, $\alpha$ and $s$ should be optimized. It is important to note that the above analysis also applies to other thin deformable objects that do not have a layered structure. In this case, eqn. (\ref{eqn:34}) should be modified to $I_{z}=\frac{b h^{3}}{12}$. From eqn. (\ref{eqn:30}), eqn. (\ref{eqn:31}), and eqn. (\ref{eqn:38}), it is evident that larger values of $m$ and $I_z$ lead to higher $F_{f2}$ values, indicating increased difficulty in successful grasping. Therefore, a book with larger mass and moment of inertia is selected as the object of analysis: $m=1kg$, $b=300mm$, $l=200mm$, $h=15mm$, $h_{\text {paper }}=0.18mm$, and $E=828MPa$ \cite{lee2016bending}. In addition, the friction coefficients of the contact surfaces are selected as $\mu_1=\mu_3=0.2$, and $\mu_2=0.6$ \cite{skedung2011tactile}.

Figure. 1 in Appendix B shows the curve of $F_{f2}$ during the book repositioning process from start to finish. It can be observed that the fingertip angle $\alpha$ and the spacing $s$ have significant effects on the force $F_{f2}$, with larger $\alpha$ and smaller $s$ leading to a considerable increase in $F_{f2}$, potentially causing $F_{f2}$ to enter the failure region. Therefore, considering structural dimensions, the design parameter is selected as $\alpha=20^{\circ}$, and the strategy parameter $s$ should be larger. According to eqn. (\ref{eqn:25}), $s$ is directly proportional to the finger opening angle $\theta_2$. The fingers should open as wide as possible within the effective range, which is approximately $40mm$.

\begin{figure*}[t]
\vspace{-0mm}
\centerline{\includegraphics[width=1.9\columnwidth]{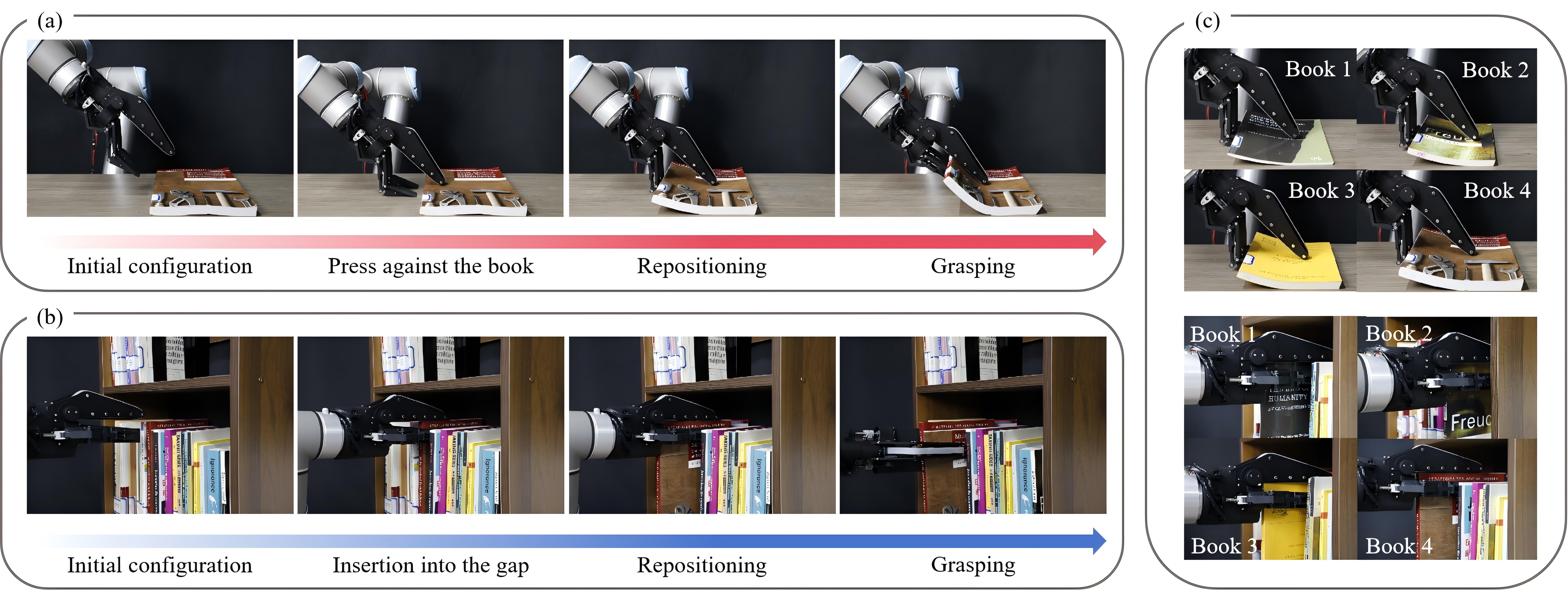}}
\vspace{-0mm}
\caption{\small Book grasping experiments. (a) Steps for grasping books on the desktop. (b) Steps for grasping books on the bookshelf. (c) Results of repositioning four books in different scenarios.}
\label{Fig:6}
\vspace{-0mm}
\end{figure*}

\section{Experiments}
\subsection{Grasping Books on the Desktop}
We designed a gripper prototype and mounted it on a UR5e robotic arm to test the grasping tasks of four books with different sizes and surface friction coefficients, specific details of the gripper prototype can be found in Appendix B. First, we tested the friction coefficient of the objects, as described in Appendix B. Based on the grasping strategy in Section V, the specific grasping steps are illustrated in Fig. \ref{Fig:6}(a): 1) In the initial state, the book is placed flat on the desktop, and the gripper is adjusted to the initial configuration; 2) The robotic arm moves downward until the active surface thumb contacts the surface of the book; 3) The belt on the thumb rotates to pull the book into the gripper's interior; 4) The fingers close to grasp the book. To make the repositioning process in step 3 approximate a quasi-static process, a low speed for the active surface was set, with the speed of motor 3 set to $166rpm$, resulting in a belt speed of $0.036 m/s$. It is worth noting that the control of the four steps mentioned above is extremely simple, with control inputs provided sequentially without any coupled motion control. Furthermore, except for step 1, which has three control inputs ($\psi$, $\theta_2$, $\theta_3$), each of the remaining steps has only one control input (which corresponds to the robotic arm's movement distance, the belt's movement distance, and $\theta_2$, respectively). We conducted 20 sets of experiments for each of the four books to test the grasping success rate. The experimental results are shown in the Table \ref{tab:tables2}, and the experimental demonstration can be seen in the video. The results indicate that the gripper can successfully grasp the thinner three books without errors, while it failed twice when grasping the thickest book, with failures occurring at the stage when the book just contacted the fingers due to its high bending stiffness. In addition, we conducted the same 20 sets of experiments from the non-bound side of the book, where the non-bound side is oriented towards the fingertips. The experimental results show that the grasping success rate for all four books is $100\%$, as seen in Fig. 3 in Appendix B. Therefore, the grasping strategy does not suffer a reduction in success rate due to the separable characteristics of the non-bound side of the book. Overall, the experimental results demonstrate that the gripper can adapt to grasping various types of books, with the thickest book having a thickness of $14 mm$ and a bending stiffness ($EI_{z}$) of $7.7e^{-3}Nm^{2}$.

\subsection{Grasping Books on the Bookshelf}
Similarly, we conducted grasping tests for books on the bookshelf. The specific steps are shown in the Fig. \ref{Fig:6}(b): 1) Adjust the gripper to the initial configuration and align it with the target book; 2) Insert the fingers into the gap between the target book and the surrounding books, while the thumb presses against the side of the book; 3) Rotate the belt on the thumb to pull the book into the gripper's interior; 4) Change the gripper's configuration and grasp the book. The speed of the active surface during the repositioning process was set to the same as in the desktop grasping task. Similarly, there is no coupled motion control in any of the four steps. The experimental results indicate that there are few instances of failure when grasping the four books, as shown in Table \ref{tab:tables2}. This is primarily due to positional deviations of the fingers under open-loop control, causing the fingers to be obstructed by the books during insertion. This error can be mitigated in the future with vision-based detection methods. However, there were no failures after successfully completing the insertion actions, demonstrating the good stability of the grasping method based on the active surface.

\begin{table}[b]
\centering
\renewcommand{\arraystretch}{1.5}
\setlength{\tabcolsep}{3pt}
\caption{Success Rate of Experimental Objects}
\label{tab:tables2}
\begin{threeparttable}
\begin{tabular}{ccc|cc}
\hline
Object & Success Rate 1 & Success Rate 2 & Object       & Success Rate 1 \\ \hline
book 1 & 20/20          & 18/20          & A4 paper     & 20/20          \\
book 2 & 20/20          & 17/20          & plastic film & 20/20          \\
book 3 & 20/20          & 18/20          & mouse pad    & 20/20          \\
book 4 & 18/20          & 16/20          & fabric       & 20/20          \\ \hline
\end{tabular}
\begin{tablenotes}[para,flushleft]
\item Success Rate 1 represents the success rate of grasping objects on the desktop, and Success Rate 2 represents the success rate of grasping objects on the bookshelf.
\end{tablenotes}
\end{threeparttable}
\end{table}

\subsection{Other Applications}
\subsubsection{Grasping Thin Objects}
We also conducted grasping experiments on four extremely thin objects placed flat on the desktop, including plastic film, fabric, A4 paper, and mouse pad. The grasping strategy is the same as in the book grasping experiments, as shown in Fig. \ref{Fig:S3}. These thin objects exhibit significant differences in characteristics, as listed in Table  III in Appendix B. Specifically, plastic film and mouse pad have a higher friction coefficient with the thumb, while A4 paper and fabric have a lower friction coefficient with the thumb. Additionally, the four objects vary in softness, with the fabric and mouse pad being softer, and the plastic film and A4 paper relatively harder. The results show that the gripper can grasp these four thin objects without errors, seen in Table \ref{tab:tables2}. Therefore, the proposed gripper can not only grasp books with significant differences in size, weight, and surface friction coefficients but also handle extremely thin deformable objects (with the thinnest thickness of $0.08 mm$), demonstrating excellent versatility.

\begin{figure}[t]
\vspace{-0mm}
\centerline{\includegraphics[width=0.92\columnwidth]{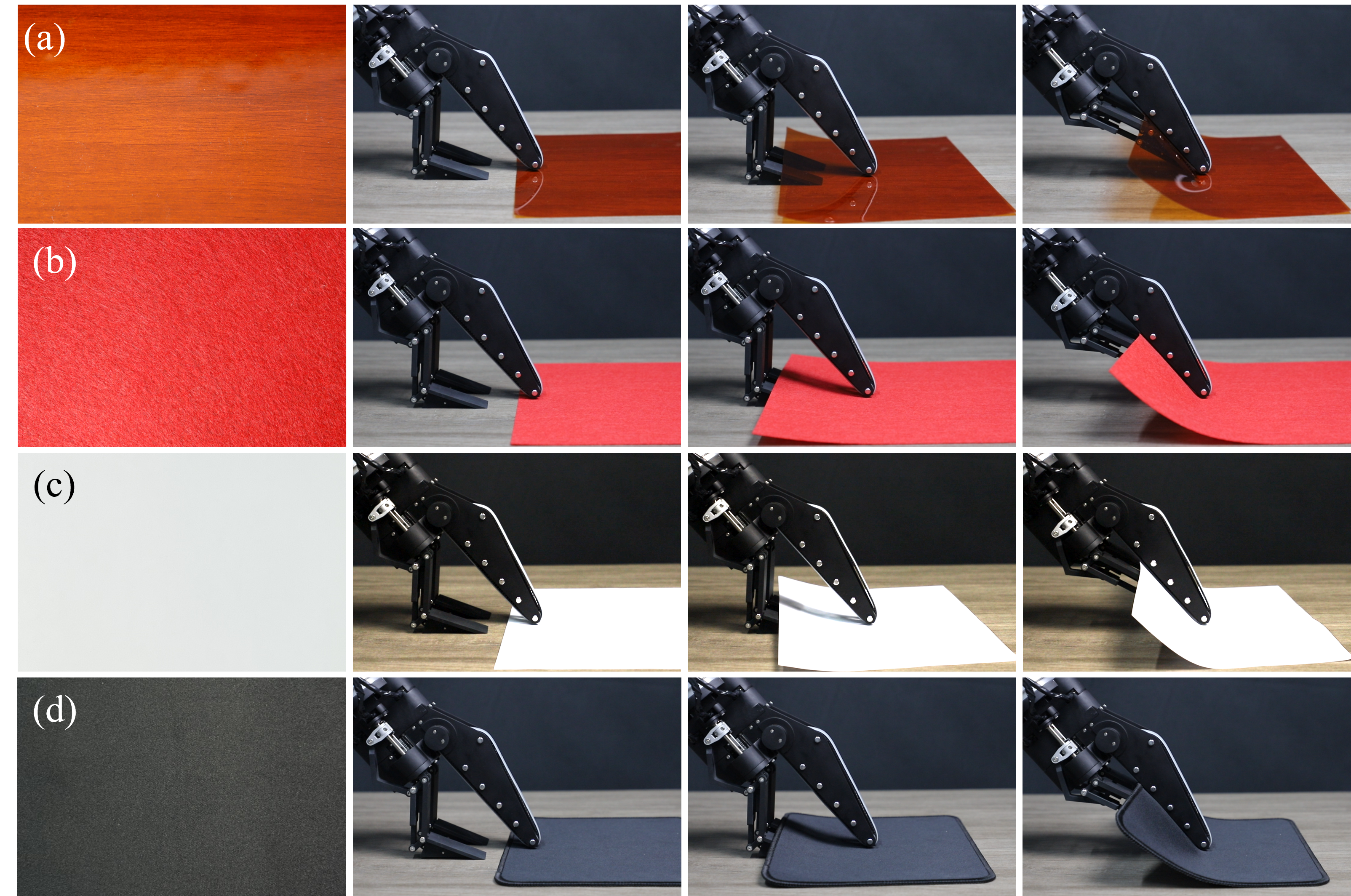}}
\vspace{-0mm}
\caption{\small Thin object grasping experiments. (a) Plastic film. (b) Fabric. (c) A4 Paper. (d) Mouse pad.}
\label{Fig:S3}
\vspace{0mm}
\end{figure}

\subsubsection{“Grasp-Place” Task}
\begin{figure}[t]
\vspace{-0mm}
\centerline{\includegraphics[width=0.92\columnwidth]{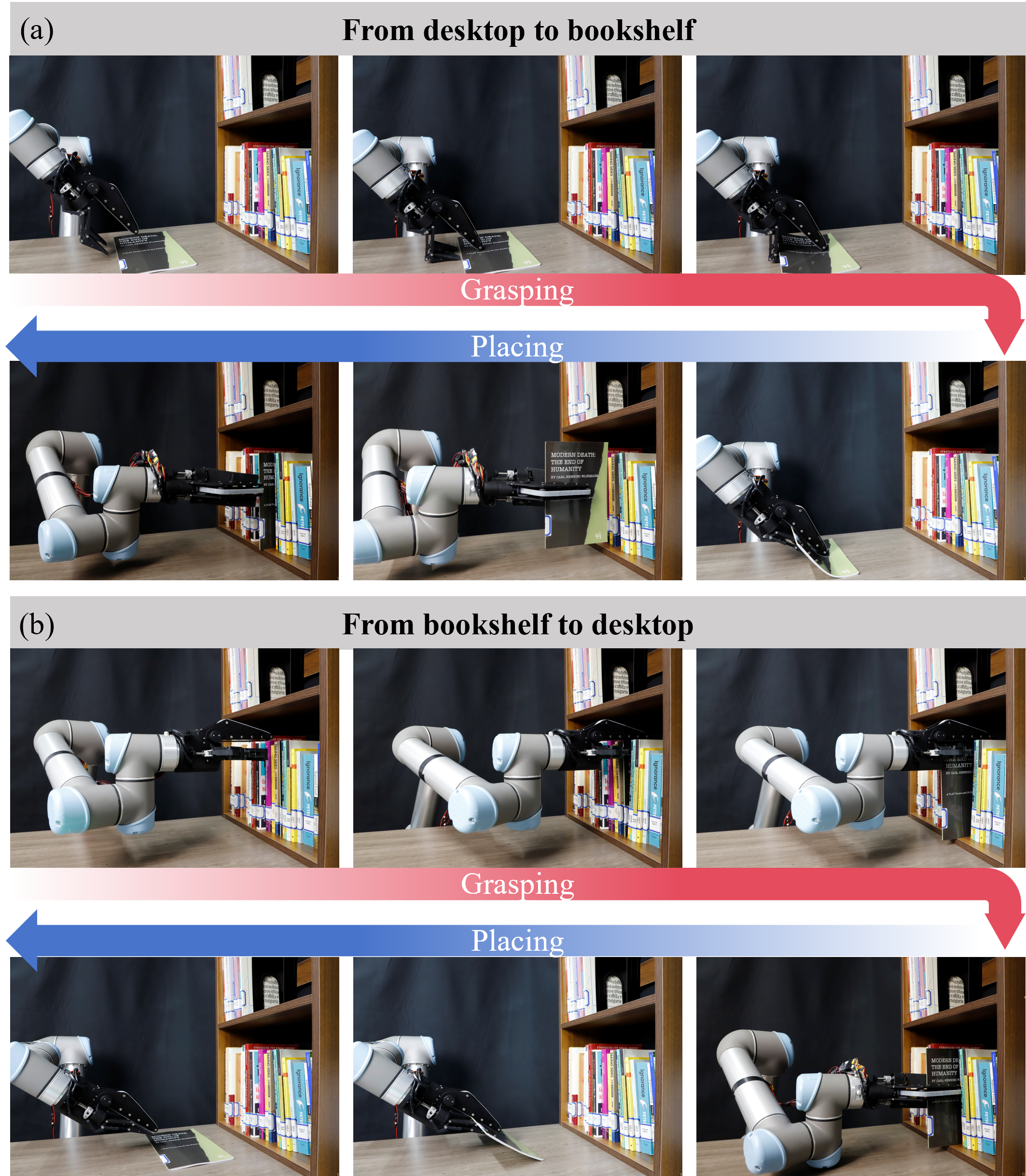}}
\vspace{-0mm}
\caption{\small “Grasp-place” task sequence. (a) Grasp a book from the desktop and place it onto the bookshelf. (b) Grasp a book from the bookshelf and place it onto the desktop.}
\label{Fig:S4}
\vspace{-0mm}
\end{figure}

As shown in Fig. \ref{Fig:S4}, we simulated a scenario of automated book management in a library. According to the proposed grasping strategy, the gripper grasped a book placed flat on the desktop and then placed it on the bookshelf, and also grasped a book from the bookshelf and placed it back on the desktop. Notably, it is very challenging for traditional grippers to place books on the bookshelf because the contact points between the fingers and the books are fixed. To completely place a book inside the bookshelf, the entire gripper needs to be moved to the inside of the bookshelf, and the finger structure may be obstructed by surrounding books, making this process difficult to achieve. In contrast, the proposed gripper does not require opening the fingers. Instead, it utilizes the active surface's reverse rotation to pull the book out while still holding it. The feasibility of this process primarily relies on the low-friction fingertip surface and the high-friction active surface. Thus, the placement stage eliminates the need for the fingers to push the book into the shelf, resulting in a more stable overall process. The specific operational workflow can be seen in the video.

\section{Conclusion} 
\label{sec:conclusion}

In this paper, we propose a novel reconfigurable gripper with an active surface and innovatively introduce an active-surface-based grasping strategy for thin deformable objects, demonstrating excellent generality. Specifically, the active surface provides the object with a stable and continuously changing contact interface, enabling repositioning of thin objects. This method avoids the complex finger gait control typically required when grasping thin objects, while the control quantities in each grasping step are input sequentially, eliminating coupled motion control. Additionally, the small-angle wedge-shaped fingertips smooth the deformation process of thin objects, thereby enhancing the robustness of the grasping process. The reconfigurable mechanism endows the gripper with greater versatility, allowing it to handle vertically arranged books. Experimental results show that the gripper achieves a high success rate when grasping books in different states as well as various thin deformable objects. This study not only provides a low-control-complexity and generalizable method for grasping thin deformable objects but also demonstrates the feasibility of replacing multi-finger dexterous hands with active surfaces in robotic grasping and manipulation.

Future work will introduce perception systems, such as vision, to achieve automated long-sequence tasks for thin deformable objects like books. Additionally, we plan to incorporate active or passive adjustment mechanisms on the active surface thumb to handle more types of thin deformable objects. We will further explore the integration of active surfaces with dexterous hands to accomplish a wider range of object grasping and manipulation tasks.

\section*{Acknowledgments}
This work was supported in part by the National Natural Science Foundation of China Youth Program under Grant No. 52305037, Zhejiang Provincial Natural Science Foundation of China under Grant No. LD26E050001, the "Pioneer" and "Leading Goose" R\&D Programs of Zhejiang Province under Grants  No. 2025C01072, the National Natural Science Foundation of China Youth Program under Grant No. 52505041, and the China Postdoctoral Science Foundation under Grant No. 2024M762814.

\newpage


\bibliographystyle{unsrt}
\bibliography{references}

\end{document}


\title{Appendix}

\maketitle

\section*{Appendix A}
The transformation matrix from the gripper coordinate system $\{O\}$ to the proximal joint coordinate system $\{A\}$ can be obtained as follows:
\begin{equation}
\label{eqn:5}
{ }^{O} \boldsymbol{T}_{A}=\left[\begin{array}{cccc}
\cos \theta_{1} & -\sin \theta_{1} & 0 & 0 \\
\sin \theta_{1} & \cos \theta_{1} & 0 & -w \\
0 & 0 & 1 & 0 \\
0 & 0 & 0 & 1
\end{array}\right]
\end{equation}

The transformation matrix from the proximal joint coordinate system $\{A\}$ to the endpoint $F$ of the proximal joint driving rod is:
\begin{equation}
\label{eqn:6}
{ }^{A} \boldsymbol{T}_{F}=\left[\begin{array}{cccc}
1 & 0 & 0 & 0 \\
0 & \cos \theta_{2} & -\sin \theta_{2} & -l_{4} \cos \theta_{2}-l_{3} \sin \theta_{2} \\
0 & \sin \theta_{2} & \cos \theta_{2} & -l_{4} \sin \theta_{2}+l_{3} \cos \theta_{2} \\
0 & 0 & 0 & 1
\end{array}\right]
\end{equation}

Therefore, the transformation matrix from the gripper coordinate system $\{O\}$ to the endpoint $F$ is the product of these two matrices:
\begin{equation}
\label{eqn:7}
\begin{array}{l}
{ }^{O} \boldsymbol{T}_{F}={ }^{O} \boldsymbol{T}_{A}{ }^{A} \boldsymbol{T}_{F} \\
=\left[\begin{array}{cccc}
\cos \theta_{1} & -\sin \theta_{1} \cos \theta_{2} & \sin \theta_{1} \sin \theta_{2} & \\
\sin \theta_{1} & \cos \theta_{1} \cos \theta_{2} & -\cos \theta_{1} \sin \theta_{2} & { }^{O} \boldsymbol{p}_{F} \\
0 & \sin \theta_{2} & \cos \theta_{2} & \\
0 & 0 & 0 & 1
\end{array}\right]
\end{array}
\end{equation}
Where ${ }^{O} \boldsymbol{p}_{F}$ represents the coordinates of the endpoint $F$ in the gripper coordinate system $\{O\}$:
\begin{equation}
\label{eqn:8}
{ }^{O} \boldsymbol{p}_{F}=\left[\begin{array}{c}
l_{4} \sin \theta_{1} \cos \theta_{2}+l_{3} \sin \theta_{1} \sin \theta_{2} \\
-l_{4} \cos \theta_{1} \cos \theta_{2}-l_{3} \cos \theta_{1} \sin \theta_{2}-w \\
-l_{4} \sin \theta_{2}+l_{3} \cos \theta_{2}
\end{array}\right]
\end{equation}

Let the distance from the origin $H$ of the slider coordinate system to $x_O$ when the gripper is in its initial configuration be $h_0$. The lead of the screw is $S$, and the rotation angle of the screw is $\alpha$. Therefore, the position of the slider at any point can be expressed as:
\begin{equation}
\label{eqn:9}
h=h_{0}+\frac{\alpha S}{2 \pi}
\end{equation}

The transformation matrix from the gripper coordinate system $\{O\}$ to the slider coordinate system $\{H\}$ is:
\begin{equation}
\label{eqn:10}
{ }^{O} \boldsymbol{T}_{H}=\left[\begin{array}{cccc}
1 & 0 & 0 & 0 \\
0 & 1 & 0 & -w \\
0 & 0 & 1 & -h \\
0 & 0 & 0 & 1
\end{array}\right]
\end{equation}

The transformation matrix from the slider coordinate system $\{H\}$ to the endpoint G of the link $GH$ is:
\begin{equation}
\label{eqn:11}
{ }^{H} \boldsymbol{T}_{G}=\left[\begin{array}{cccc}
\cos \theta_{3} & -\sin \theta_{3} & 0 & l_{6} \sin \theta_{3} \\
\sin \theta_{3} & \cos \theta_{3} & 0 & -l_{6} \cos \theta_{3} \\
0 & 0 & 1 & 0 \\
0 & 0 & 0 & 1
\end{array}\right]
\end{equation}

Therefore, the transformation matrix from the gripper coordinate system $\{O\}$ to the endpoint G is:
\begin{equation}
\label{eqn:12}
{ }^{O} \boldsymbol{T}_{G}={ }^{O} \boldsymbol{T}_{H}{ }^{H} \boldsymbol{T}_{G}=\left[\begin{array}{cccc}
\cos \theta_{3} & -\sin \theta_{3} & 0  \\
\sin \theta_{3} & \cos \theta_{3} & 0 & { }^{O} \boldsymbol{p}_{G} \\
0 & 0 & 1 & \\
0 & 0 & 0 & 1
\end{array}\right]
\end{equation}
where ${ }^{O} \boldsymbol{p}_{G}$ represents the coordinates of the endpoint $G$ in the gripper coordinate system $\{O\}$:
\begin{equation}
\label{eqn:13}
{ }^{O} \boldsymbol{p}_{G}=\left[\begin{array}{c}
l_{6} \sin \theta_{3} \\
-l_{6} \cos \theta_{3}-w \\
-h
\end{array}\right]
\end{equation}

\begin{figure*}[t]
\vspace{-0mm}
\centerline{\includegraphics[width=2\columnwidth]{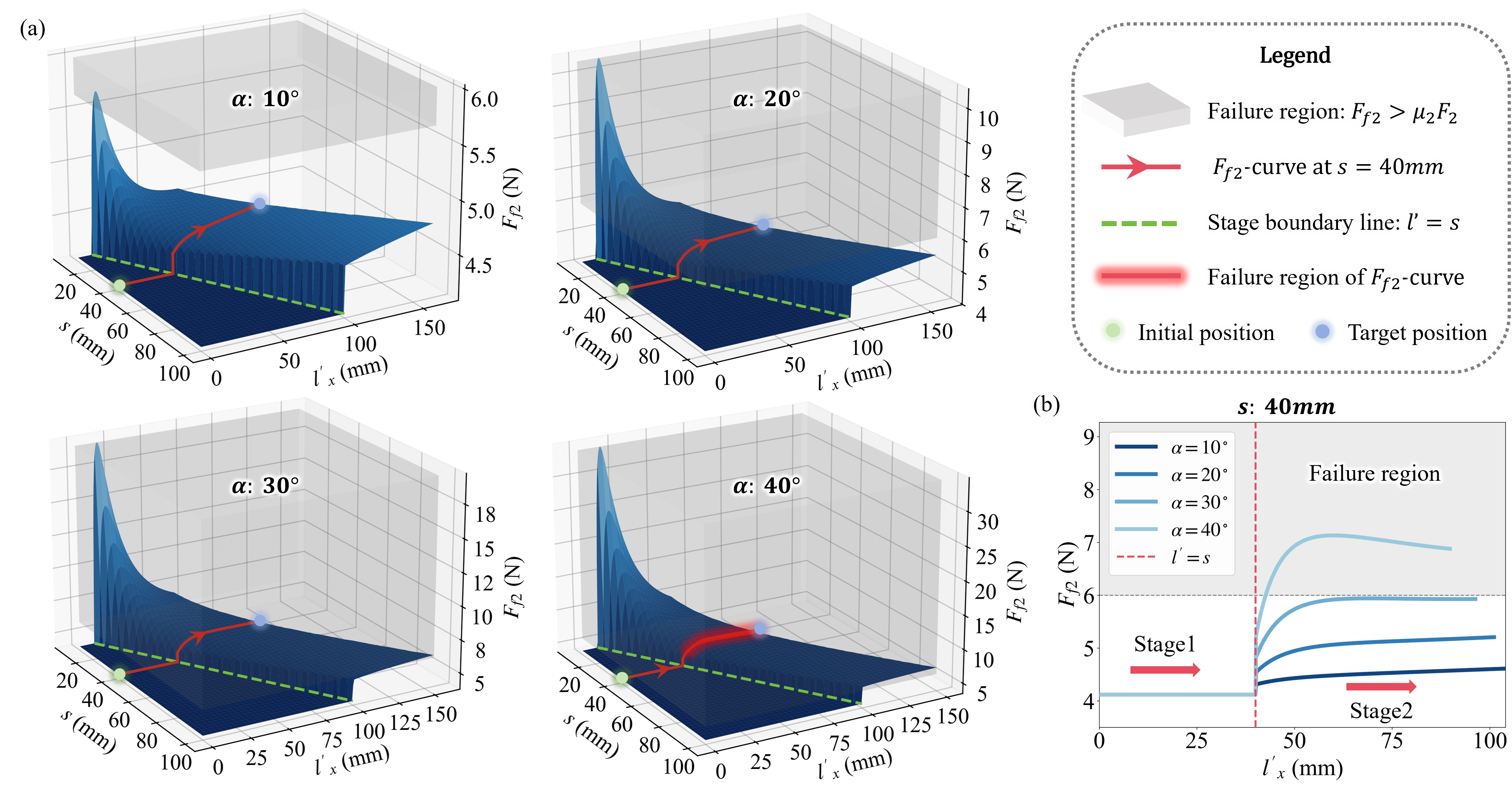}}
\vspace{-0mm}
\caption{\small $F_{f2}$-curves under different parameters. (a) $F_{f2}$ for different distances $s$ under four fingertip angles $\alpha$. (b) $F_{f2}$-curves for four fingertip angles when $s=40mm$.}
\label{Fig:S1}
\vspace{-0mm}
\end{figure*}

As the mechanism AEFGH is a planar crank-slider mechanism, it has the following constraints:
\begin{equation}
\label{eqn:14}
\left\{\begin{array}{c}
\left\|{ }^{O} \boldsymbol{p}_{F}-{ }^{O} \boldsymbol{p}_{G}\right\|=l_{5} \\
\theta_{3}=\theta_{1}
\end{array}\right.
\end{equation}

Therefore, we can obtain the equations involving $h$ and $\theta_{2}$:
\begin{equation}
\label{eqn:15}
2\left(h l_{4}+l_{3} l_{6}\right) \sin \theta_{2}-2\left(h l_{3}-l_{4} l_{6}\right) \cos \theta_{2}=l_{3}^{2}+l_{4}^{2}-l_{5}^{2}+l_{6}^{2}+h^{2}
\end{equation}

Thus
\begin{equation}
\label{eqn:16}
\cos \left(\theta_{2}-\phi\right)=\frac{C}{\sqrt{A^{2}+B^{2}}}
\end{equation}
where $\cos \phi=\frac{A}{\sqrt{A^{2}+B^{2}}}$, $A=2\left(h l_{3}-l_{4} l_{6}\right)$, $\quad B=-2\left(h l_{4}+l_{3} l_{6}\right)$, and $\quad C=l_{5}^{2}-l_{3}^{2}-l_{4}^{2}-l_{6}^{2}-h^{2}$.

With the above equations, once the input quantity $h$ is determined, the angle $\theta_2$ of finger opening and closing can be calculated.

The transformation matrix from the proximal joint coordinate system $\{A\}$ to the distal joint coordinate system $\{B\}$ can be express as:
\begin{equation}
\label{eqn:17}
{ }^{A} \boldsymbol{T}_{B}=\left[\begin{array}{cccc}
1 & 0 & 0 & 0 \\
0 & 1 & 0 & -l_{1} \sin \theta_{2} \\
0 & 0 & 1 & l_{1} \cos \theta_{2} \\
0 & 0 & 0 & 1
\end{array}\right]
\end{equation}

The transformation matrix from the distal joint coordinate system $\{B\}$ to the endpoint P is:
\begin{equation}
\label{eqn:18}
{ }^{B} \boldsymbol{T}_{P}=\left[\begin{array}{cccc}
1 & 0 & 0 & 0 \\
0 & \cos \theta_{4} & \sin \theta_{4} & l_{7} \sin \theta_{4} \\
0 & -\sin \theta_{4} & \cos \theta_{4} & l_{7} \cos \theta_{4} \\
0 & 0 & 0 & 1
\end{array}\right]
\end{equation}

The coordinate transformation from the gripper coordinate system $\{O\}$ to the endpoint P can be expressed as follows:
\begin{equation}
\label{eqn:19}
\begin{array}{l}
{ }^{O} \boldsymbol{T}_{P}={ }^{O} \boldsymbol{T}_{A}{}^{A} \boldsymbol{T}_{B}{}^{B} \boldsymbol{T}_{P} \\
=\left[\begin{array}{cccc}
\cos \theta_{1} & -\sin \theta_{1} \cos \theta_{4} & -\sin \theta_{1} \sin \theta_{4} & \\
\sin \theta_{1} & \cos \theta_{1} \cos \theta_{4} & \cos \theta_{1} \sin \theta_{4} & { }^{O} \boldsymbol{p}_{P} \\
0 & -\sin \theta_{4} & \cos \theta_{4} & \\
0 & 0 & 0 & 1
\end{array}\right]
\end{array}
\end{equation}
where ${ }^{O} \boldsymbol{p}_{P}$ represents the coordinates of the endpoint P in the gripper coordinate system:
\begin{equation}
\label{eqn:20}
{ }^{O} \boldsymbol{p}_{P}=\left[\begin{array}{c}
l_{1} \sin \theta_{1} \sin \theta_{2}-l_{7} \sin \theta_{1} \sin \theta_{4} \\
l_{7} \cos \theta_{1} \sin \theta_{4}-l_{1} \cos \theta_{1} \sin \theta_{2}-w \\
l_{7} \cos \theta_{4}+l_{1} \cos \theta_{2}
\end{array}\right]
\end{equation}

Similarly, the coordinates of the other finger endpoint Q in the gripper coordinate system can be obtained as:
\begin{equation}
\label{eqn:21}
{ }^{O} \boldsymbol{p}_{Q}=\left[\begin{array}{c}
l_{1} \sin \theta_{1} \sin \theta_{2}-l_{7} \sin \theta_{1} \sin \theta_{4} \\
-l_{7} \cos \theta_{1} \sin \theta_{4}+l_{1} \cos \theta_{1} \sin \theta_{2}+w \\
l_{7} \cos \theta_{4}+l_{1} \cos \theta_{2}
\end{array}\right]
\end{equation}

Based on the eqn. (\ref{eqn:20}) and eqn. (\ref{eqn:21}), once $h$, $\theta_1$, $\theta_2$, and $\theta_4$ are determined, the positions of the endpoints of the two fingers can be obtained.

\vspace{10mm}
\section*{Appendix B}
\subsection*{B-1 Parameter Optimization}

\vspace{-0mm}
\begin{table}[h]
\centering
\renewcommand{\arraystretch}{1.5}
\caption{Control and Object Parameters}
\vspace{-0mm}
\label{tab:tables0}
\centering
\begin{tabular}{>{\centering\arraybackslash}p{0.9cm}p{2.7cm}|>{\centering\arraybackslash}p{0.9cm}p{2.7cm}}
\hline
\multicolumn{1}{>{\centering\arraybackslash}p{0.9cm}}{Parameter} &
\multicolumn{1}{>{\centering\arraybackslash}p{2.7cm}|}{Description} &
\multicolumn{1}{>{\centering\arraybackslash}p{0.9cm}}{Parameter} &
\multicolumn{1}{>{\centering\arraybackslash}p{2.7cm}}{Description} \\ \hline
$\theta_{1,3}$ & Finger rotation angle & $l$                             & Object length               \\
$\theta_2$             & Finger opening angle        & $l^{\prime}$                    & Length in contact point \\
$\theta_4$             & Fingertip rotation angle    & $m$                             & Object mass                 \\
$\psi$                 & Gripper rotation angle      & $\mu_{1,2,3}$ & Coefficient of friction     \\
$s$                    & Finger opening distance     & $F_{1}$                         & Finger support force        \\
$h$                    & Object thickness            & $F_{2}$                         & Thumb normal force          \\
$b$                    & Object width                & $F_{3}$                         & Desktop support force       \\ \hline
\end{tabular}
\end{table}
\vspace{-0mm}

We set the normal force applied by the thumb to the book as 10N. Based on the equation $g\left(F_{1}\right) \leq g\left(F_{2}\right)$, we plotted the curve of $F_{f2}$, as shown in Fig. \ref{Fig:S1}. It can be observed that for four different values of $\alpha$, when the value of $s$ is small, i.e., $s \in(0,30)$, the maximum value of $F_{f2}$ rapidly decreases as $s$ increases. However, for larger values of $s$, as $s$ increases, the rate of decrease in the maximum value of $F_{f2}$ becomes more gradual. This indicates that the initial value of $s$ chosen for the configuration should avoid the range of small values, so that the book repositioning process is less likely to enter a failure region. On the other hand, the magnitude of the fingertip angle $\alpha$ also significantly affects $F_{f2}$, with a smaller $\alpha$ implying a smaller maximum value of $F_{f2}$ during the grasping process, thus avoiding entry into the failure zone. Therefore, a smaller fingertip angle should be chosen in the structural design.

\subsection*{B-2 Gripper Prototype}
We fabricated a prototype of the gripper, with the housing of the gripper completed through 3D printing. The main transmission structures, including the linkages and fingertips, were manufactured using aluminum. The active surface on the thumb consists of an S3M timing belt with a 1mm thick layer of silicone strip attached to its surface. Silicone adhesive was used to bond the two ends of the strip together while maintaining surface flatness. Three motors (DJI, M2006 P36) were placed on the base of the gripper. Benefiting from the modular design, the base, fingers, and thumb of the gripper can be assembled separately and then joined together. The main structural parameters of the gripper are shown in the Table \ref{tab:tables3}.

\begin{table}[h]
\centering
\renewcommand{\arraystretch}{1.5}
\caption{Design Parameters}
\label{tab:tables3}
\begin{tabular}{cc|cc}
\hline
Parameter & Value  & Parameter & Value        \\ \hline
$l_1$        & 60mm   & $l_7$        & 65mm         \\
$l_2$        & 15mm   & $l_8$        & 125mm        \\
$l_3$        & 15mm   & $w$         & 27mm         \\
$l_4$        & 15mm   & $e$         & 37mm         \\
$l_5$        & 62.5mm & $a$         & 20°          \\
$l_6$        & 15mm   & $h$         & (39mm, 63mm) \\ \hline
\end{tabular}
\end{table}

\subsection*{B-3 Friction Coefficient Acquisition}

\begin{table}[b]
\centering
\renewcommand{\arraystretch}{1.5}
\caption{Parameters of Experimental Objects}
\label{tab:tables1}
\begin{tabular}{cccccc}
\hline
Object       & Size (mm)    & Mass (g) & u1   & u2   & u3   \\ \hline
book 1       & 155×229×4    & 89       & 0.22 & 0.75 & 0.18 \\
book 2       & 129×198×9    & 131      & 0.30 & 0.50 & 0.18 \\
book 3       & 150×220×10   & 240      & 0.39 & 0.57 & 0.20 \\
book 4       & 175×246×14   & 440      & 0.26 & 0.98 & 0.21 \\
A4 paper     & 210×297×0.10 & 4        & 0.31 & 0.53 & 0.18 \\
plastic film & 210×320×0.08 & 4        & 0.24 & 0.82 & 0.17 \\
mouse pad    & 200×221×2.6  & 22       & 0.39 & 0.72 & 0.31 \\
fabric       & 330×330×0.61 & 10       & 0.25 & 0.51 & 0.19 \\ \hline
\end{tabular}
\end{table}

\begin{figure}[h]
\vspace{-0mm}
\centerline{\includegraphics[width=0.95\columnwidth]{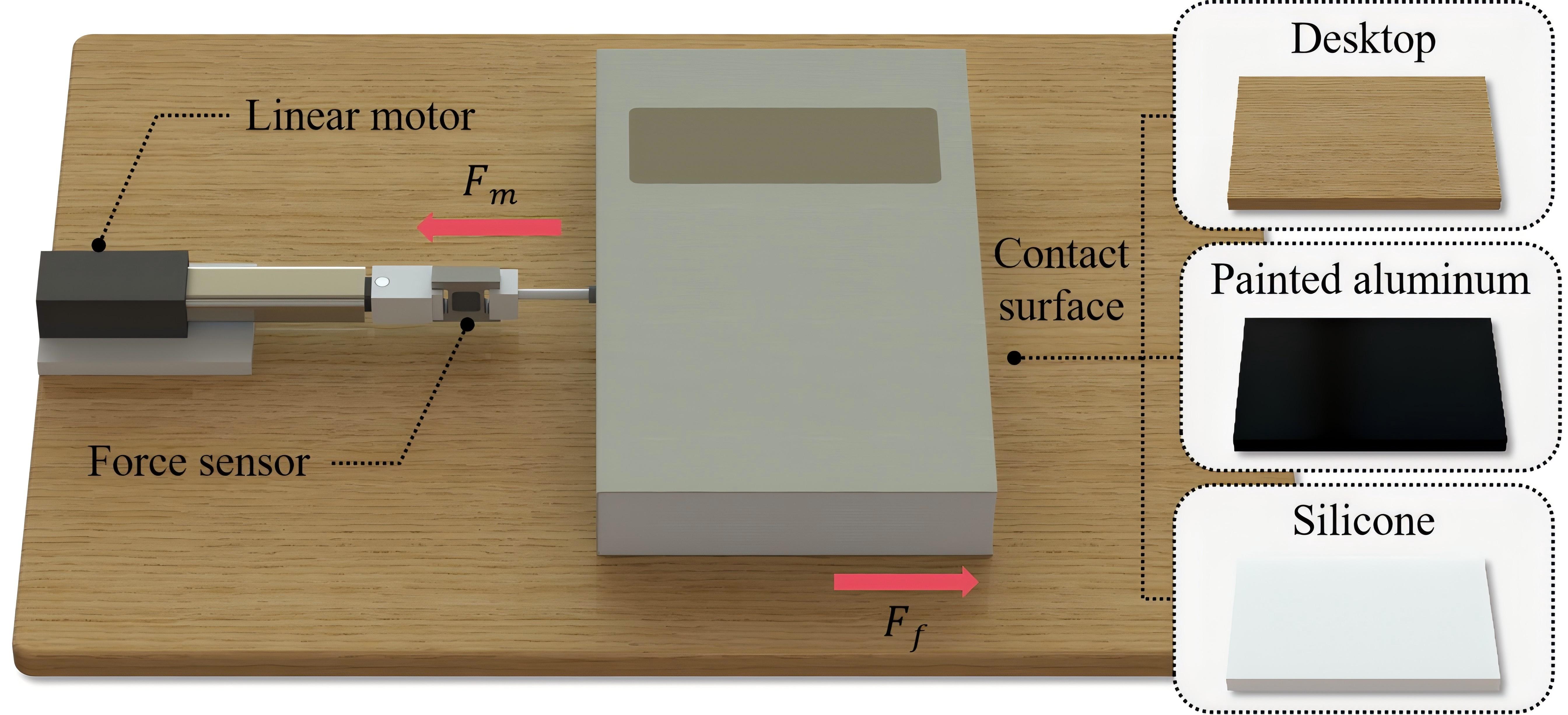}}
\vspace{-0mm}
\caption{\small Friction coefficient testing platform for objects.}
\label{Fig:S2}
\vspace{0mm}
\end{figure}

We set up an experimental platform to obtain the friction coefficients of books and other thin objects, as shown in Fig. \ref{Fig:S2}. Specifically, a linear actuator (Hoodland, IP60YR) was fixed on the desktop, with a force sensor (Simbatouch, SBT630) installed on the output rod of the motor. The other end of the force sensor was connected to the object, which was placed flat on three different surfaces. The motor pulled the object at a constant speed, and the force sensor recorded the steady pulling force $F_m$. The friction coefficient was then calculated using the formula $\mu=F_{m} / m g$, where $m$ is the mass of the object. The results are shown in Table \ref{tab:tables1}.

\subsection*{B-4 Grasping from the Non-Bound Side of the Book}
\begin{figure}[h]
\vspace{-3mm}
\centerline{\includegraphics[width=1\columnwidth]{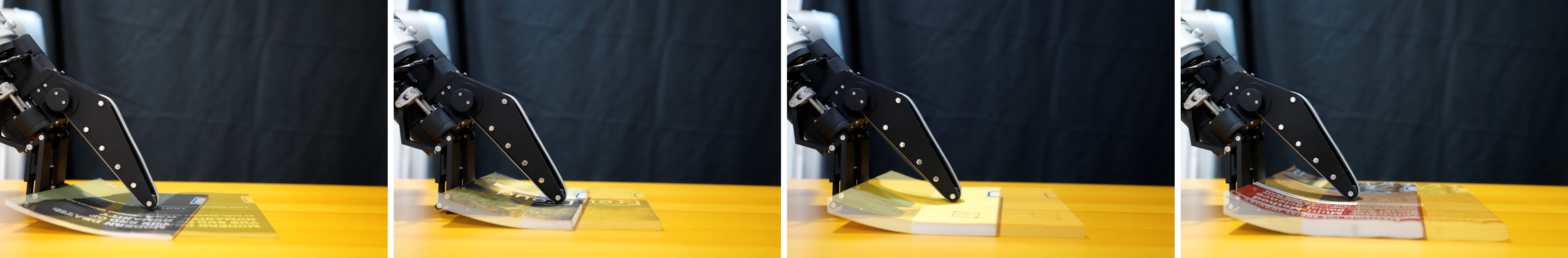}}
\vspace{-1mm}
\caption{\small Grasping experiments from the non-bound side of the book.}
\label{Fig:r1}
\vspace{-2mm}
\end{figure}

For books placed flat on a desktop, we conducted grasping experiments from the non-bound side of the book, where the non-bound side was oriented towards the fingers. Unlike grasping from the bound side, the pages on the non-bound side are separable, making it easier for interlayer sliding to occur during the grasping process. Additionally, after contact with the fingers, only some pages may successfully undergo subsequent repositioning operations. The same grasping strategy was used for books placed flat on the desktop, and 20 grasping trials were performed on four different books, as shown in the Fig. \ref{Fig:r1}. The grasping success rate for all four books was $100\%$. This success is attributed to the small-angle wedge-shaped fingertip design, which allows for easy separation of extremely thin objects from the table during the grasping process.

\IEEEpeerreviewmaketitle



\bibliographystyle{plainnat}